\documentclass{article}
\usepackage{ijcai24}

\usepackage{times}
\usepackage{soul}
\usepackage{url}
\usepackage[hidelinks]{hyperref}
\usepackage[utf8]{inputenc}
\usepackage[small]{caption}
\usepackage{graphicx}
\usepackage{amsmath}
\usepackage{amsthm}
\usepackage{booktabs}
\usepackage{algorithm}
\usepackage[noend]{algpseudocode}

\usepackage[switch]{lineno}

\usepackage{multirow}
\usepackage{amsfonts}
\usepackage{subcaption}
\usepackage{amsmath}
\usepackage{xcolor}
\usepackage{amssymb}

\title{MARCO: A Memory-Augmented Reinforcement Framework \\ for Combinatorial Optimization\footnote{This is an extended version of the paper accepted at the Thirty-Third International Joint Conference on Artificial Intelligence (IJCAI) 2024. The original version is available at https://doi.org/10.24963/ijcai.2024/766.}}

\author{
Andoni I. Garmendia$^1$
\and
Quentin Cappart$^2$\and
Josu Ceberio$^1$\And
Alexander Mendiburu$^{1}$
\affiliations
$^1$University of the Basque Country (UPV/EHU), Donostia-San Sebastian, Spain\\
$^2$Polytechnique Montréal, Montreal, Canada\\
\emails
\{andoni.irazusta, josu.ceberio, alexander.mendiburu\}@ehu.eus,
quentin.cappart@polymtl.ca
}

\begin{document}

\maketitle

\begin{abstract}
\textit{Neural Combinatorial Optimization} (NCO) is an emerging domain where deep learning techniques are employed to address combinatorial optimization problems as a standalone solver. Despite their potential, existing NCO methods often suffer from inefficient search space exploration, frequently leading to local optima entrapment or redundant exploration of previously visited states. This paper introduces a versatile framework, referred to as \textit{Memory-Augmented Reinforcement for Combinatorial Optimization} (MARCO), that can be used to enhance both constructive and improvement methods in NCO through an innovative memory module. MARCO stores data collected throughout the optimization trajectory and retrieves contextually relevant information at each state. 
This way, the search is guided by two competing criteria: making the best decision in terms of the quality of the solution and avoiding revisiting already explored solutions. This approach promotes a more efficient use of the available optimization budget. Moreover, thanks to the parallel nature of NCO models, several search threads can run simultaneously, all sharing the same memory module, enabling an efficient collaborative exploration.
Empirical evaluations, carried out on the maximum cut, maximum independent set and travelling salesman problems, reveal that the memory module effectively increases the exploration, enabling the model to discover diverse, higher-quality solutions. MARCO achieves good performance in a low computational cost, establishing a promising new direction in the field of NCO.
\end{abstract}



\section{Introduction}
\label{intro}

The objective in Combinatorial Optimization (CO) problems is to find the optimal solution from a finite or countable infinite set of discrete choices. These problems are prevalent in many real-world applications, such as chip design~\cite{mirhoseini2021graph}, genome reconstruction~\cite{vrvcek2022learning} and program execution~\cite{gagrani2022neural}. 

In recent years, the field of \textit{Neural Combinatorial Optimization} (NCO) has emerged as an alternative tool for solving such problems~\cite{bengio2021machine,mazyavkina2021reinforcement,bello2016neural}. NCO uses deep neural networks to address CO problems in an end-to-end manner, learning from data and generalizing to new, unseen instances. Researchers in this field have followed the steps of heuristic optimization, proposing the neural counterparts of \textit{constructive methods}~\cite{bello2016neural,kool2018attention,kwon2020pomo} and \textit{improvement methods}~\cite{lu2019learning,chen2019learning,wu2021learning}.

\textit{Neural constructive methods} quickly generate an approximate solution in a one-shot manner by means of a learnt neural model. While being simple and direct, constructive methods suffer from their irreversible nature, barring the possibility of revisiting earlier decisions. This limitation becomes particularly pronounced in large problems where suboptimal initial decisions in the construction of the solution can significantly impact the final outcome. To improve the performance of these methods, recent efforts have employed techniques such as \textit{sampling}, where instead of following the output of the model deterministically, a random sample is taken from a probability distribution given by the output, with the intention of obtaining better solutions and break with the deterministic behaviour, obtaining a richer set of solutions; or \textit{beam search}~\cite{choo2022simulation}, which maintains a collection of the highest-quality solutions as it explores the search space based on the output of the neural network, i.e., the probability of adding an item to the partial solution that is being constructed. Similarly, \textit{active search}~\cite{bello2016neural,hottung2021efficient} is used to update the model's weights (or a particular set of weights) during test time, in order to overfit the model to the test instance to be solved.

Alternatively, \textit{neural improvement methods} are closely linked to perturbation methods, such as local search. They start from a complete solution, and operate by iteratively suggesting a modification that improves the current solution at the present state. 
Unlike constructive methods, improvement methods inherently possess the ability to explore the search space of complete solutions. However, they often get stuck in local optima or revisit the same states repeatedly, leading to cyclical patterns. Recent studies~\cite{barrett2020exploratory,garmendia2023neural} have employed a variety of strategies inherited from the combinatorial optimization literature to tackle these drawbacks. The method by~\cite{barrett2020exploratory} keeps a record of previously performed actions, while the study in~\cite{garmendia2023neural} maintains a tabu memory of previously visited states, forbidding the actions that would lead to visit those states again.

Neural constructive methods, neural improvement methods, and most classical optimization proposals all face a significant challenge: \textit{exploring efficiently the search space}. To address this, we introduce a new framework, referred to as 
\textit{Memory-Augmented Reinforcement for Combinatorial Optimization}, or MARCO. This framework integrates a memory module into both neural constructive and neural improvement methods.
The memory records the visited or created solutions during the optimization process, and retrieves relevant historical data directly into the NCO model, enabling it to make more informed decisions. 

A key feature of MARCO is the ability to manage a shared memory when several search \textit{threads} are run in parallel. 
By doing so, MARCO not only reduces the redundancy of storing similar data across multiple threads but also facilitates a collaborative exploration of the search space, where each thread benefits from a collective understanding of the instance.

The main contributions of the paper are as follows: \textbf{(1)} introducing MARCO as a pioneering effort in integrating memory modules within both neural improvement and constructive methods. \textbf{(2)} Designing a similarity-based search mechanism that retrieves past, relevant information to feed the memory and to better inform the model. \textbf{(3)} Presenting the parallelism capabilities of MARCO, which enables a more efficient and collaborative exploration process. \textbf{(4)} Illustrating the implementation of the framework to three graph-based problems: maximum cut, maximum independent set, and travelling salesman problem. Experiments are then carried out on these three problems with graphs up to 1200 nodes. The empirical results indicate that MARCO surpasses some of the recently proposed learning-based approaches, demonstrating the benefits of using information regarding visited solutions. The source code and supplementary material are available online\footnote{https://github.com/TheLeprechaun25/MARCO.}.

\section{Related Work}

Various strategies have been developed to enhance the exploration of the search space in NCO algorithms. Most of the methods sample from the model's logits~\cite{bello2016neural,kool2018attention,kwon2020pomo}, which introduces stochasticity into the solution inference process. Beyond sampling, entropy regularization has been implemented during the training of NCO models~\cite{kim2021learning}, to ensure the models are not overconfident in their output. Furthermore, \cite{grinsztajn2024winner} proposed a multi-decoder system, where each decoder is trained on instances where it performs best, resulting in a set of specialized and complementary policies.


Despite these advancements, none of these methods exploit any kind of memory mechanism, which has the potential to leverage previous experiences in the decision-making process and promote exploration.


In the work by \cite{garmendia2023neural}, a \textit{tabu search} algorithm \cite{glover1993user}, known for its memory-based approach to circumventing cyclical search patterns, is layered on top of a neural improvement method. The algorithm utilizes a tabu memory to track previously visited solutions. However, this memory serves merely as an external filter, preventing the selection of tabu actions without integrating historical data into the neural model's decision-making process.

DeepACO \cite{ye2023deepaco} uses a neural network to learn the underlying heuristic of an \textit{ant colony optimization}  algorithm \cite{blum2005ant,dorigo2006ant}. It maintains an external pheromone matrix, indicative of promising variable decisions. However, the integration of this pheromone data is indirect; it is combined in a post-hoc fashion with the output probabilities of the model rather than being an intrinsic part of the learning process. 

Closer to our work, ECO-DQN \cite{barrett2020exploratory} is a neural improvement method that records the last occurrence of each action. This operation-based memory approach, which simply tracks when actions were last taken, is computationally efficient, requiring only minimal storage. 
The drawback of this approach is that it only focuses on the actions, failing to consider the overall search context. The effectiveness of an action is often contingent on the broader state of the optimization process, a fact that operation-based memory fails to capture. Compared to this work, we save entire solutions in memory, incorporating a more holistic view of the search context to the system, at the cost of higher memory requirements.


\section{MARCO: A Memory-Based Framework}
\label{marco}

\begin{figure*}
\centering
\includegraphics[width=0.8\linewidth]{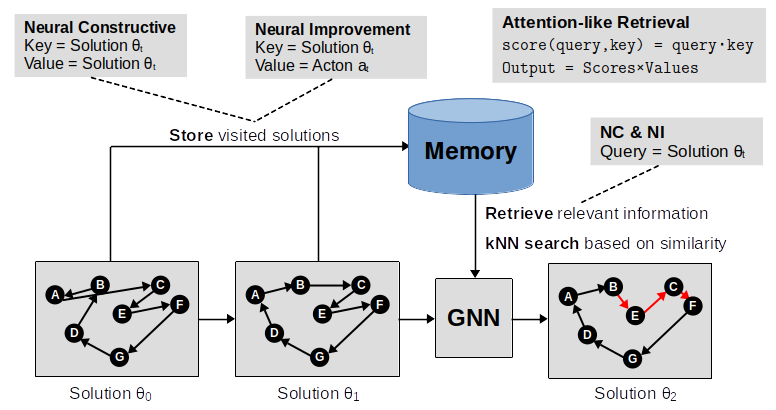} 
\caption{Schematic of the MARCO framework, illustrating its memory integration process. Each visited solution, $\sigma_t$, is stored in the memory module. During each iteration of the search process, MARCO performs a similarity-based retrieval to access relevant context from past visited solutions, retrieving the $k$ nearest solutions. 
Then, the retrieved solutions are aggregated using a weighted average, with the similarity being the weight.
The retrieval process can be seen as an attention mechanism where the current state serves as the query (Q), and stored past solutions function as keys (K) and NC methods use the same solutions as values (V), while NI methods use corresponding actions.}
\label{fig:general_marco}
\end{figure*}


This section introduces MARCO, the main contribution of this paper.
Although the framework can be used for arbitrary CO problems, we first focus on 
\textit{graph-based problems}, as they are ubiquitous in combinatorial optimization.
In fact, from the 21 NP-complete problems identified by Karp (1972), ten are decision versions of graph optimization problems, while most of the other ones can also be modeled over graphs. 

Let $G=(V,E)$ be a simple graph composed of a set of nodes $V$ and a set of edges $E$.
Finding a solution $\theta$  for graph problems often involves finding subsets of nodes or edges that satisfy specific criteria, such as minimizing or maximizing a certain objective function. 

Briefly, the idea of MARCO is to leverage both (1) a learnt policy $\pi$ 
defining how the current solution should be modified for exploring the search space,
and (2) a memory module $\mathcal{M}$, providing information to build the policy.
The policy is typically parameterized with a neural network, and especially
with a \textit{graph neural network}~\cite{kipf2016semi} when operating on graph problems. Such an architecture has been considered as highly relevant for combinatorial optimization~\cite{cappart2023combinatorial}.
Besides, the policy is iteratively called to modify the solution until a convergence threshold has been reached.

This mechanism can be integrated into both constructive and improvement methods. 
The main difference relates to \textit{how a solution is defined} and \textit{how information is retrieved from the memory}.
Let $\theta_t$ refer to a complete solution obtained after $t$ iterations, and $\hat{\theta}_t$ refer to a partial solution, i.e., a solution where only $t$ variables has been assigned, with at least one variable not assigned.
In \textit{constructive methods}, MARCO is capable of using a deterministic policy repeatedly, i.e., opting for the greedy action to generate multiple different constructions. Each construction starts from an empty solution and each optimization step consists in extending the current partial solution, i.e., assigning an unassigned variable in the optimization problem.
The policy takes as input both static information (i.e., the graph instance $G$)
and dynamic information related to the current partial solution $\hat{\theta_t}$).
The memory then stores a solution once it is completed, i.e., once all the variables have been assigned.
On the other hand, \textit{improvement methods} feature an
iterative refinement of a complete solution. Each step modifies a current (complete) solution $\theta_t$, transitioning it to a subsequent solution $\theta_{t+1}$. In this scenario, the dynamic information is the complete solution $\theta_t$.
Each explored solution is recorded into the memory.

For both methods, the training is conducted through reinforcement learning. Each time a completed solution is reached, a reward $r_t$ is obtained, denoting how good the executed optimization trajectory has been. The reward is designed to balance two factors: (1) the quality of the solution found and (2) the dissimilarity of the new solution compared to previous solutions stored in memory.

A general overview of MARCO's framework is illustrated in Figure~\ref{fig:general_marco}.

\subsection{Memory Module}

As shown in Algorithm \ref{alg_marco}, the inference in MARCO starts with the selection of an initial solution (refer to line 2). In each optimization step, the memory module $\mathcal{M}$ is responsible for storing the visited solutions (line 6), and retrieving aggregated historical data $h_t$ (line 7). The historical data ($h_t$) is aggregated with the current (partial) solution and the graph features ($G$) to form the current optimization state $s_t$ (line 8). Subsequently, $s_t$ is input into the model (line 9), which then proposes a set of actions that generate new solutions (line 10).

The specific process of retrieval is shown in Algorithm \ref{alg_retrieve}. To retrieve relevant solutions, MARCO employs a \textit{similarity-based search}. This involves comparing the current (partial) solution ($\hat{\theta_t}$ or $\theta_t$) with each stored solution ($\theta_{t'}$ where $t'<t$) using a similarity metric (e.g., the inner product in line 4). Intuitively, the idea is to fed the policy with the most similar solutions to the current one
for executing the next exploration step. We carry out the retrieval using a \textit{k-nearest neighbors search} (line 5). Rather than simply averaging the $k$ most similar solutions, MARCO uses a weighted average approach, where the weight given to each past solution is directly proportional to its similarity to the current solution. This score is normalized, ranging from 0 (completely different) to 1 (identical), to represent the level of similarity (see line 6).



\begin{algorithm}
\caption{Inference with MARCO}
\begin{algorithmic}[1]
\Procedure{MARCO}{$\text{graph} \; G, \text{policy} \; \pi, k, max\_steps$}       
\State $\theta_0 \gets \Call{InitializeSolutions}{}$
\State $\mathcal{M} \gets \Call{InitializeMemory}{k}$
\For{$t = 0$ to $max\_steps-1$}
    \If{$\theta_{t}$ is Completed}
    \State $\mathcal{M} \gets \Call{StoreInMem}{\theta_{t}}$
    \EndIf
    \State $h_t \gets \Call{RetrieveFromMem}{\mathcal{M}, k, \theta_t}$

    \State $s_t \gets \Call{Aggregate}{G, \theta_t, h_t}$ \Comment{Get current state}
    \State $a_t \gets \Call{Policy}{\pi, s_t}$
    
    \State $\theta_{t+1} \gets \Call{Step}{\theta_t, a_t}$ \Comment{Get next solution}

\EndFor
\EndProcedure
\end{algorithmic}
\label{alg_marco}
\end{algorithm}

\begin{algorithm}
\caption{Action retrieval from memory}
\begin{algorithmic}[1]
\Procedure{RetrieveFromMem}{$\mathcal{M}, k,  \theta_t$}  
    \State $v \gets \Call{Length}{\mathcal{M}}$ 
    \State $k \gets \min(k, v)$ 
    \State $simScore \gets \Call{InnerProduct}{\theta_t, \theta_{t'} \; | \; t'<t.}$ 
    \State $h_t \gets \Call{KNN}{k, simScore, \mathcal{M}}$ 
    \State $\hat{h}_t \gets h_t \times \Call{Norm}{simScore}$ 
    \State \textbf{return} $\hat{h}_t$ \Comment{Return relevant historical data}
\EndProcedure
\end{algorithmic}
\label{alg_retrieve}
\end{algorithm}

\subsubsection{Collaborative Memory}

An additional feature enabled by MARCO is the implementation of parallel optimization threads during its inference phase. In this setup, multiple concurrent threads are run for each problem instance, collaboratively exploring the solution space. A key aspect of this functionality is the use of shared memory across all threads. This collective memory stores all the explored solutions by any thread, making it accessible to the entire group.

\section{Application of MARCO}
In this study, we demonstrate the adaptability of MARCO to various problem types, encompassing both constructive and improvement methods. We specifically apply MARCO in two scenarios: (1) a neural improvement method for problems with binary variables, such as the Maximum Cut (MC) and the Maximum Independent Set problem (MIS); and (2) a neural constructive method for permutation problems, such as the Travelling Salesman Problem (TSP).

\begin{figure}
\centering
\includegraphics[width=0.98\columnwidth]{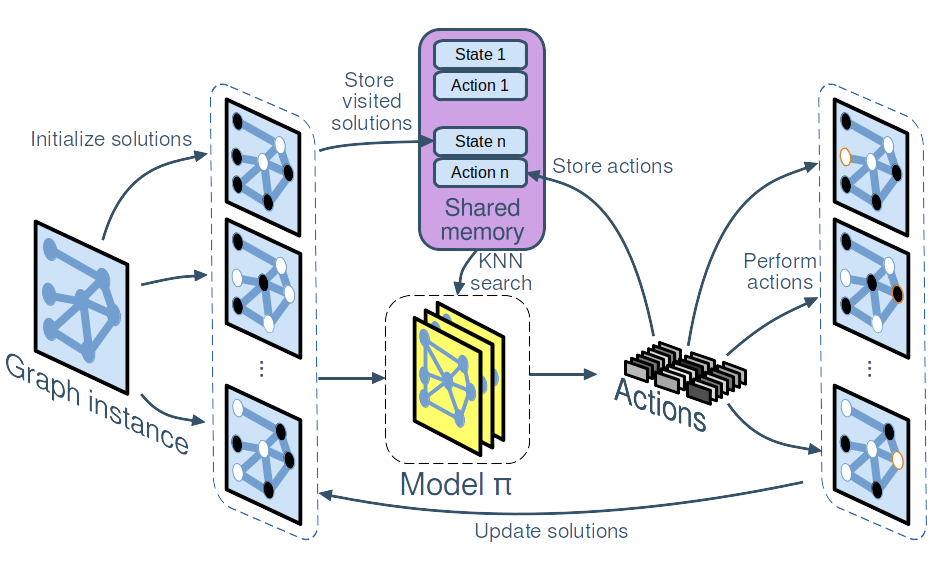} 
\caption{MARCO for Neural Improvement methods for the Maximum Cut problem. Initially, multiple solutions are randomly generated for a given problem instance. Each solution is iteratively improved, forming a thread. Throughout this process, the visited solutions and corresponding actions are stored into a shared memory. This collective memory then updates the graph features fed to the model.}
\label{fig:inference_NI}
\end{figure}

\subsection{Improvement Methods for Binary Problems}

In binary optimization problems, a solution is formalized as a binary vector, denoted as $\theta \in \{0, 1\}^{|V|}$ for a problem with $|V|$ variables. Each variable $x_i$ represents a binary decision  for the $i^{th}$ variable.
Neural improvement methods are designed to optimize a problem by iteratively refining an initial complete solution $\theta_0$, which can be generated either randomly or through heuristic methods. In the case of binary problems, the central operation is a node-wise operator that flips the value of a node in $\theta$. 
The memory records visited solutions and their associated actions (e.g., a \textit{bit-flip} action, consisting in flipping the value of a variable).
When a new solution is generated, the model consults the memory to retrieve the actions performed in similar previous scenarios. The importance of the stored actions is given by the similarity between the current solution $\theta_t$ and previously stored solutions $\theta_{t'}$ with $t'<t$. We compute the similarity using the inner product:\begin{equation}
\label{eq:sim}
     \text{Similarity}(\theta_t, \theta_{t'}) = \langle \theta_t, \theta_{t'} \rangle = \sum_{i \in V} (\theta_t)_i \cdot (\theta_{t'})_i
\end{equation}
In this case, the aggregated memory data ($h_t$) is a vector of size $|V|$, defined as the weighted average of the actions that were executed in the $k$ most similar solutions (if any). See Figure \ref{fig:inference_NI} for a visual description of the inference in neural improvement methods with MARCO.

The \textit{reward} $r_t$ obtained by a neural improvement model at each step $t$ is defined as the non-negative difference between the current objective value of the solution, $f(\theta_{t})$, and the best objective value found thus far ($f(\theta^*)$), i.e., $r_t = \max\{f(\theta_{t}) - f(\theta^*), 0\}$. This reward structure, prevalent in neural improvement methods \cite{ma2021learning,wu2021learning}, motivates the model to continually seek better solutions.
To prevent the model from cycling through the same states and encourage novel solution exploration, we incorporate a binary penalty term $p_{r}$, activated when revisiting previously encountered solutions. The adjusted reward for each step is thus $\hat{r_t} = r_t - w_p \times p_{r}$, where $w_p$ is a weight factor for the penalty.

\subsection{Constructive Methods for Permutations}

\begin{figure}
\centering
\includegraphics[width=0.9\columnwidth]{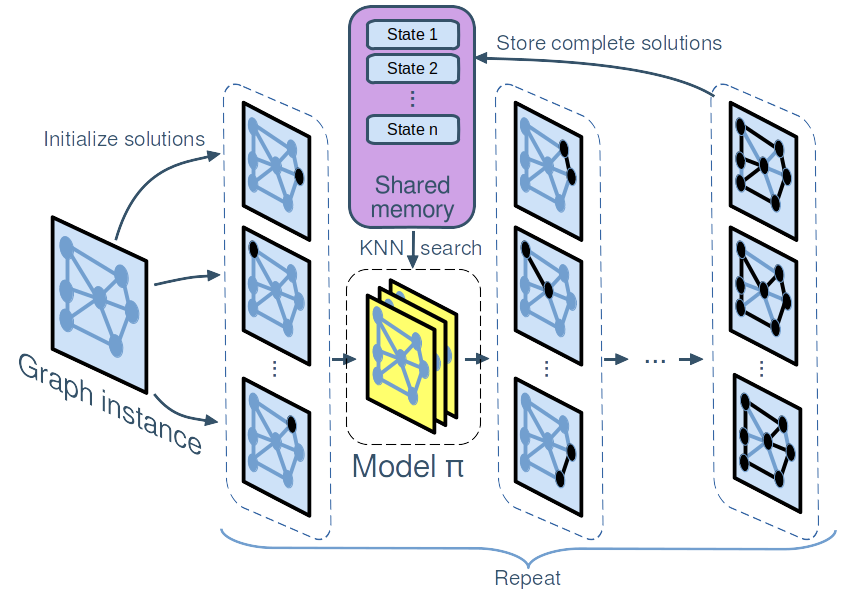} 
\caption{MARCO for Neural Constructive methods. Travelling Salesman example. Each solution in a batch begins with a distinct initial node. Subsequently, every thread proceeds to iteratively construct a solution, considering data gathered from the memory module. Upon the completion of this construction process, the obtained solution is stored within the memory, serving as a reference for subsequent solution constructions.
}
\label{fig:inference_NC}
\end{figure}

The objective in permutation problems like the TSP is to find a permutation of nodes in a graph that maximizes or minimizes a specific objective function. Neural constructive methods build the permutation incrementally, starting from an empty solution and adding elements sequentially until a complete permutation is formed. 

In the context of permutation problems, the solution $\theta$ can also be conceptualized as a binary vector $\theta^b \in \{0, 1\}^{|E|}$. Each element in this vector corresponds to an edge $e_{ij}$ in the graph, indicating whether the edge is part of the solution, that is, whether node $i$ and node $j$ are adjacent in the permutation.

At each step of the permutation building process, the model operates on a partial solution, defined as a sequence $\hat{\theta}_t^b = (\theta^b_t[1], \theta^b_t[2], \ldots, \theta^b_t[k])$, where $k < |V|$ is the current number of nodes in the sequence. As the model progresses through constructing the permutation, the memory data is used to consider which edges have been selected in previously constructed solutions that are similar to $\hat{\theta}_t^b$. 
The similarity score is performed by an inner product between the binary representations of the partial solution $\hat{\theta}_t^b$  and the complete solutions saved in memory $\theta^b_{t'}$ with $t'<t$:
\begin{equation}
\label{eq:sim2}
     \text{Similarity}(\hat{\theta}^b_t, \theta^b_{t'}) = \langle \hat{\theta}^b_t, \theta^b_{t'} \rangle = \sum_{i \in V} (\hat{\theta}^b_t)_i \cdot (\theta^b_{t'})_i
\end{equation}
Figure \ref{fig:inference_NC} showcases the inference in MARCO for neural constructive methods.
Training involves computing a \textit{reward} once the solution is completed. The reward $r_t = f(\theta_{t})$, given by the objective value of the solution, is adjusted by subtracting a baseline value to stabilize training. A common approach is to use the average reward across different initializations, as done in POMO for the TSP~\cite{kwon2020pomo}.

Our initial experiments with constructive models showed that exact solution repetitions are uncommon for large instances. Therefore, instead of the binary penalty system used in improvement methods, we apply a scaled penalization based on similarity levels with stored solutions. The final reward is calculated as $\hat{r_t} = r_t - w_p \times \Call{AvgSim}{\theta_t, \theta_{t'}}$, where $\Call{AvgSim}{\theta_t, \theta_{t'}}$ is the average of all the inner products between the constructed solution and the $k$ most similar stored solutions.


\section{Model Architecture}
Graph neural networks are particularly well-suited to parameterize policy $\pi$. We specifically use a Graph Transformer (GT)~\cite{dwivedi2020generalization} coupled with a Feed-Forward Neural Network. GTs are a generalization of transformers~\cite{vaswani2017attention} for graphs. The fundamental operation in GTs involves applying a shared self-attention mechanism in a fully connected graph, allowing each node to gather information from every other node. The gathered information is then weighted by computed attention weights, which indicate the importance of each neighbor's features to the corresponding node.

Our model aims to be adaptable to various combinatorial problems, 
requiring it to assimilate both the graph's structural information and the attributes of its nodes and edges.
To achieve this, we modify the GT to incorporate structural information encoded as edge features within the Attention (Attn) mechanism. This adaptation is reflected in the following equation.
\begin{equation}
   \text{Attn}(Q, K, V) = \left(\text{softmax}\left(\frac{QK^T}{\sqrt{d_k}} + \mathbf{E}\right) \cdot \mathbf{E}\right)V 
\end{equation}
In this equation, $Q$, $K$, and $V$ stand for Query, Key, and Value, respectively, which are fundamental components of the attention mechanism~\cite{vaswani2017attention} and $d_k$ is a scaling factor. $\mathbf{E} = \mathbf{W}_{e} \cdot \mathbf{e}_{ij}$ is a linear transformation of the edge weights, where $\mathbf{W}_e \in \mathbb{R}^{1 \times n_{\text{heads}}}$ is a learnable weight matrix, and $\mathbf{e}_{ij}$ represents the edge features between nodes $i$ and $j$. $\mathbf{E}$  integrates edge information by being added to the attention scores and used in a dot product.

The final step involves processing the output of the GT through an element-wise feed-forward neural network to generate action probabilities. The output of the GT could be both node- or edge-embeddings. The performed action depends on the method in use. In our studied cases, we will use node embeddings to generate node-based actions: node-flips for improvement methods in binary problems and node addition to the partial solution for the constructive method in permutation problems. However, MARCO is also applicable to edge-based actions, such as pairwise operators (swap, 2-opt) for permutation-based improvement methods.

We utilize the policy gradient REINFORCE algorithm \cite{williams1992simple} to find the optimal parameter set, $\pi^*$, which maximizes the expected cumulative reward in the optimization process.

\section{Experiments}

\subsection{Problems}

We validate the effectiveness of MARCO across a diverse set of CO problems both binary and permutation-based: the Maximum Cut (MC), Maximum Independent Set (MIS) and the Travelling Salesman Problem (TSP).

\paragraph{Maximum Cut (MC).} The objective in MC~\cite{dunning2018works} is to partition the set of nodes \( V \) in a graph $G$ into two disjoint subsets \( V_1 \) and \( V_2 \) such that the number of edges between these subsets (the cut) is maximized. The objective function can be expressed as:
\(
\max \sum_{(u, v) \in E} \delta_{[\theta_u \neq \theta_v]}
\)
where \( \theta_u \) and \( \theta_v \) are binary variables indicating the subset to which nodes \( u \) and \( v \) belong, and $\delta$ is a function which equals to 1 if \( \theta_u \) and \( \theta_v \) are different and 0 otherwise.

\paragraph{Maximum Independent Set (MIS).} For the MIS problem~\cite{lawler1980generating}, the goal is to find a binary vector $\theta$ that represents a subset of nodes \( S \subseteq V \) in a graph $G$ such that no two nodes in \( S \) are adjacent, and the size of \( S \) is maximized. The objective function can be formulated as:
\(
\max |S| \text{ such that } (u, v) \notin E \text{ for all } u, v \in S
\)


\paragraph{Travelling Salesman Problem (TSP).} In TSP~\cite{lawler1986traveling,wang2021deep}, given a set of nodes \( V \) and distances \( d_{u, v} \) between each pair of nodes \( u, v \in V \), the task is to find a permutation \( \theta \) of nodes in \( V \) that minimizes the total travel distance. This is expressed as:
\(
\min \sum_{i=1}^{|V|} d(\theta_i, \theta_{i+1}) \text{ with } \theta_{|V| + 1} = \theta_1
\)



\subsection{Experimental Setup}

\paragraph{Training} For each problem, we train a unique model, using instances that vary in size. This helps the model to learn strategies that can be transferable between differently sized instances. For the MC and MIS, we used randomly generated Erdos-Renyi (ER)~\cite{erdHos1960evolution} graphs with 15\% of edge probability, and sizes ranging from 50 to 200 nodes. For the TSP, fully connected graphs ranging from 50 to 100 nodes were generated, in which cities were sampled uniformly in a unit square. 
The total training time depends on the problem. The models for both MC and MIS required less than 40 minutes, while the one for the TSP required a significantly longer training (4 days) to reach convergence. 
See the supplementary material for a detailed description of the training configuration used. 

\paragraph{Inference} To evaluate the performance of MARCO, we have established certain inference parameters. For MC and MIS, we set the neural improvement methods to execute with 50 parallel threads (processing 50 solutions simultaneously), stopping upon $2|V|$ improvement steps. For the TSP, we use 100 parallel initializations (as done in POMO~\cite{kwon2020pomo}) and 20 iterations (solution constructions) for each instance. We have used $k=20$ for the similarity search. A more detailed description of the inference configuration used is reported in the supplementary material.
MARCO has been implemented using \textit{PyTorch 2.0}. A \textit{Nvidia A100} GPU has been used to train the models and perform inference. 
Exact methods and heuristics serving as baselines were executed in a cluster with \textit{Intel Xeon X5650} CPUs. 

\paragraph{Evaluation Data} Following the experimental setup of recent works~\cite{ahn2020learning,bother2021s,zhang2023let}, we will evaluate the MC and MIS problems in ER graphs of sizes between 700-800, and harder graphs from the RB benchmark~\cite{xu2000exact} of sizes between 200-300 and 800-1200.
For TSP, we follow the setting from~\cite{kool2018attention} and use randomly generated instances, with uniformly sampled cities in the unit square. We use graphs of sizes 100, 200 and 500.

\paragraph{Ablations}
We evaluate MARCO through several ablations that help us understand the impact of its different components. We begin by evaluating standalone models proposed in this work: the Neural Improvement Method (NIM) and Neural Constructive Methods (NCM), both of which operate without any integrated memory module. Next, for improvement methods, we add a NIM equipped with an operation-based memory (Op-NIM), tracking the number of steps since each action was executed lastly (imitating ECO-DQN~\cite{barrett2020exploratory}). Finally, we asses MARCO-ind, a variant of MARCO that operates without shared memory, executing multiple threads simultaneously but \textit{ind}ependently, with each thread maintaining its own separate memory.

\paragraph{Baselines}
To assess MARCO's performance, we conduct a comprehensive comparison against a broad spectrum of combinatorial optimization methods tailored to each specific problem addressed. Our comparative analysis includes exact algorithms, heuristics, and learning-based approaches

For the MC, our comparison includes the GUROBI solver~\cite{gurobi}, the local search enhanced heuristic BURER~\cite{burer2002rank}, and ECO-DQN~\cite{barrett2020exploratory}, which is a neural improvement method incorporating an operation-based memory.

For MIS, we also include GUROBI~\cite{gurobi}, together with K{\scriptsize A}MIS~\cite{lamm2016finding}, a specialized algorithm for MIS; and a constructive heuristic (Greedy), that selects the node with minimum degree in each step. Furthermore, we examine also a range of recently proposed learning-based methods: DGL~\cite{bother2021s}, LwD~\cite{ahn2020learning} and FlowNet~\cite{zhang2023let}.

For the TSP, we report results of the well known conventional solver Concorde~\cite{applegate2006concorde}, the heuristic LKH-3~\cite{papadimitriou1992complexity}, the Nearest Neighbor (NN) heuristic; and the learning-based methods used are the neural constructive POMO~\cite{kwon2020pomo} enhanced with sampling and data augmentation, LEHD~\cite{luo2024neural} which reports the best results among neural methods and two of the state-of-the-art neural improvement methods: DACT~\cite{ma2021learning} and NeuOPT~\cite{ma2023learning}. 

\subsection{Ablation Study}
\label{main_results}

\begin{figure*}[t]
    \centering
    \begin{subfigure}{0.48\linewidth}
        \includegraphics[width=\linewidth]{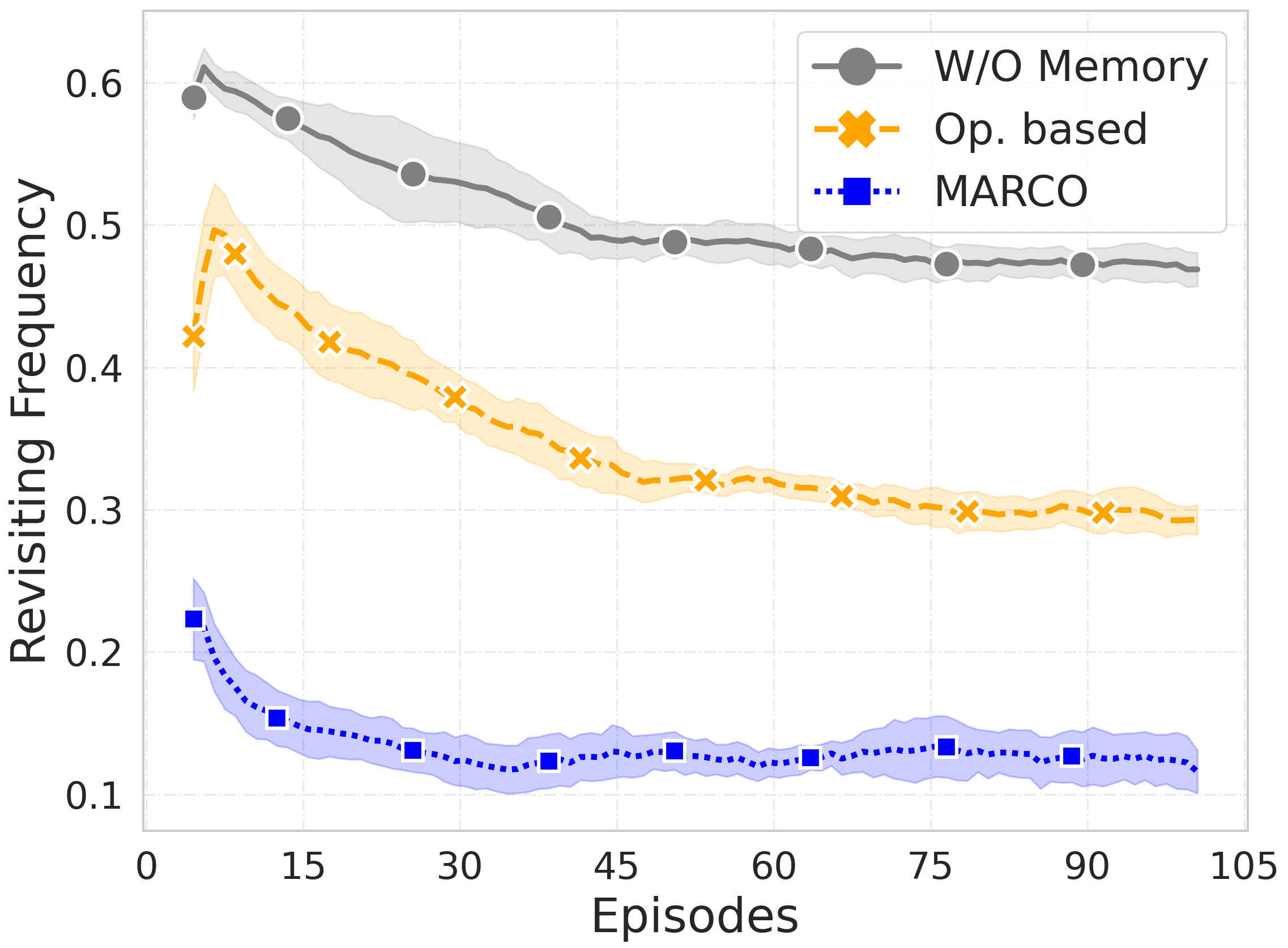}
        \caption{Revisit frequency}
    \end{subfigure}
    \hfill 
    \begin{subfigure}{0.48\linewidth}
        \includegraphics[width=\linewidth]{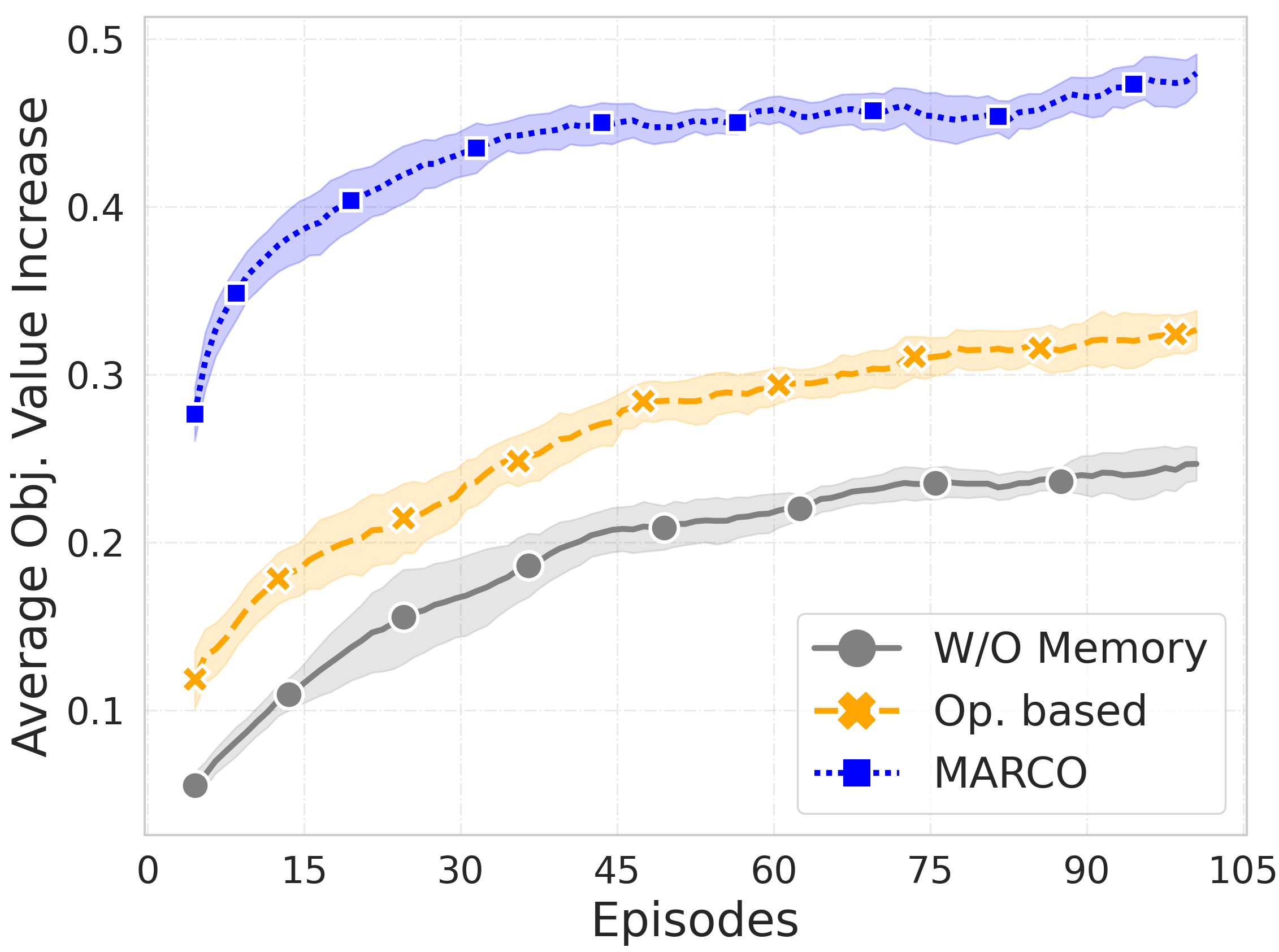}
        \caption{Reward of the model}
    \end{subfigure}
    \caption{Ablation Study}
    \label{fig:ablation}
\end{figure*}

We evaluate the impact of the proposed memory module on our model's learning dynamics and performance. Figure \ref{fig:ablation} (a) illustrates the revisit frequency of already visited solutions during the first 100 training episodes. This metric reflects the model's ability to explore diverse solutions effectively. Figure \ref{fig:ablation} (b) shows the increase in the objective value of solutions as training progresses.

The results indicate that the model without memory tend to repeat actions more frequently. In contrast, incorporating an operation-based memory module reduces this repetitiveness, facilitating broader exploration. Furthermore, MARCO's memory module demonstrates superior exploration capabilities among the tested configurations.

Overall, the study confirms that integrating the proposed memory module not only improves the model's exploration capabilities but also leads to a measurable improvement in the quality of the generated solutions.

\subsection{Performance Results}
We present the results for each studied problem in a table divided by three row-segments, the first one consisting of non-learning methods (exact and heuristic), the second with recent learning methods from the literature, and the third with the methods (MARCO and ablations) proposed in this paper.
We report both the average objective value in the evaluation instance set and the time needed for performing inference with a unique instance (batch size of 1). We use \textit{ms}, \textit{s} and \textit{m} to denote milliseconds, seconds and minutes, respectively.
For learning methods, we report the results from the best performing configuration reported in the original paper. For exact solvers, we report the best found solution when the optimal solution is not achieved in a limit of 1 and 10 minutes per instance.

\paragraph{MC}
In Table \ref{table:mc_perf} we report the results for the MC. MARCO significantly outperforms GUROBI and ECO-DQN, especially in larger problem instances (ER700-800, RB800-1200). In addition, MARCO proves to be competitive against the state-of-the-art heuristic, BURER, in the studied graph instances. The ablation results show that using the proposed memory scheme is superior to (1) not using any memory module, and (2) using an operation-based memory. Moreover, using a shared memory slightly improves the performance (with respect to MARCO-ind), while the computational cost is reduced. Compared to the ECO-DQN in computational cost, MARCO reduces the time needed to perform $2|V|$ improvement steps.

\renewcommand{\arraystretch}{1.1}
\setlength{\tabcolsep}{.30em}
\begin{table}[!tbh]
\centering 
\small
\begin{tabular}{@{}lcrcrcr@{}} 
\toprule
   & \multicolumn{2}{c}{\textbf{ER700-800}} & \multicolumn{2}{c}{\textbf{RB200-300}} & \multicolumn{2}{c}{\textbf{RB800-1200}}   \\
  \textbf{Method} & Obj. $\uparrow$ & Time & Obj. $\uparrow$ & Time & Obj. $\uparrow$ & Time \\
\midrule
GUROBI & 23420.17 & 1m & 2024.55 & 1m & 20290.08 &  1m    \\
GUROBI$_{long}$ & 24048.93 & 10m & 2286.48 & 10m & 23729.44 &  10m    \\
BURER  & \textbf{24235.93} & 1.0m & \textbf{2519.47} & 1.0m & \textbf{29791.52} & 1.0m  \\
\midrule
 ECO-DQN & 24114.06 & 2.1m & 2518.76 & 29s & 29638.78 & 3.0m \\
 \midrule
 NIM  & 24037.66 & 45s & 2517.01 & 1.5s & 29752.92 & 2.0m \\
 Op-NIM  & 24081.18 & 47s & 2518.34 & 1.6s & 29751.87 & 2.1m   \\
 MARCO-ind  & 24203.11 & 52s & 2519.46 & 2.3s  & 29778.84 & 2.7m  \\
 MARCO & \textbf{24205.97} & 49s & \textbf{2519.47} & 2.2s & \textbf{29780.71} & 2.5m \\
\bottomrule
\end{tabular}
\caption{MC performance table. The best results overall and the best results among learning-based methods are highlighted in bold.}
\label{table:mc_perf}
\end{table}

\paragraph{MIS}
Table \ref{table:mis_perf} summarizes the results for MIS. Here, MARCO is also able to surpass the learning methods and its ablations, obtaining a comparable performance to the exact solver. Moreover, it reduces the gap to the specialized K{\scriptsize A}MIS algorithm. While incorporating a memory module in MARCO (NIM vs. MARCO) increases the time cost, it contributes to achieving superior solutions, while NIM gets stuck in suboptimal solutions (increasing the number of steps does not increase the performance).

\renewcommand{\arraystretch}{1.1}
\setlength{\tabcolsep}{.50em}
\begin{table}[!tbh]
\centering 
\small
\begin{tabular}{@{}lcrcrcr@{}} 
\toprule
   & \multicolumn{2}{c}{\textbf{ER700-800}} & \multicolumn{2}{c}{\textbf{RB200-300}} & \multicolumn{2}{c}{\textbf{RB800-1200}}   \\
  \textbf{Method} & Obj. $\uparrow$ & Time & Obj. $\uparrow$ & Time & Obj. $\uparrow$ & Time \\
\midrule
GUROBI  & 43.47 & 1m & 19.98 & 1m & 40.90 & 1m \\
GUROBI$_{long}$  & 43.64 & 10m & 20.03 & 10m & 41.34 & 10m \\
 K{\scriptsize A}MIS  & \textbf{44.98} & 1m & \textbf{20.10} & 1m & \textbf{43.15} & 1m  \\
Greedy  & 38.85 & 50ms & 18.41 & 4ms & 37.78 & 54ms  \\
\midrule
DGL  & 38.71 & 11s & 19.01 & 2s & 32.32 & 3s  \\
LwD  & 41.17 & 4s & 17.36 & 1s & 34.50 & 1s  \\
FlowNet  & 41.14 & 2s &  19.18 & 0.1s & 37.48 & 0.5s \\
\midrule
NIM  & 40.16 & 2s &  19.26 & 0.5s & 37.80 &  1s  \\
Op-NIM & 40.66 & 4s &   19.70 & 1.2s & 38.59 & 4s \\
MARCO-ind  & 43.72 & 19s & 19.77 & 1.5s & 39.94 & 7s  \\
MARCO & \textbf{43.78} & 17s & \textbf{19.87} & 1.4s & \textbf{40.13} &  6s  \\
\bottomrule
\end{tabular}
\caption{MIS performance table. The best results overall and the best results among learning-based methods are highlighted in bold.}
\label{table:mis_perf}
\end{table}

\paragraph{TSP}
Results for the TSP are reported in Table \ref{table:tsp_perf}. MARCO can obtain good inference performance in the studied instances, reaching the best found solutions for N100 and N200; and being second on N500, only surpassed by LEHD. It is important to note that our basic NCM implementation (without memory) obtains comparable results with the state-of-the-art learning method while being orders of magnitude faster. Also, MARCO improves over both NCM and the method without sharing memory (MARCO-ind).

\renewcommand{\arraystretch}{1.1}
\setlength{\tabcolsep}{.50em}
\begin{table}[!tbh]
\centering 
\small
\begin{tabular}{@{}lcrcrcr@{}} 
\toprule
   & \multicolumn{2}{c}{\textbf{N100}} & \multicolumn{2}{c}{\textbf{N200}} & \multicolumn{2}{c}{\textbf{N500}}   \\
  \textbf{Method} & Obj. $\downarrow$ & Time & Obj. $\downarrow$ & Time & Obj. $\downarrow$ & Time \\
\midrule
Concorde  & \textbf{7.76} & 1m & \textbf{10.72} & 1m & \textbf{16.59} &  1m \\
LKH-3  & \textbf{7.76} & 1m & \textbf{10.72} & 1m & \textbf{16.59} & 1m  \\
NN   & 9.69 & 1ms & 13.45 & 2ms & 20.80 &  5ms \\
\midrule
POMO   & 7.81 & 0.1s & 11.73 & 1s & 21.88 & 2s  \\
LEHD  & \textbf{7.76} & 2m & \textbf{10.72} & 4m & \textbf{16.63} & 10m \\
DACT   & 7.77 & 3m & 14.23 & 5m & 145.78 & 11m \\
NeuOPT & 7.77 & 2m & 10.73 & 4m & 39.19 & 8m  \\
\midrule
NCM   & \textbf{7.76} & 0.1s &  10.75 & 1s &  16.90 & 2s  \\
MARCO-ind   & \textbf{7.76} & 3s & 10.73 & 11s & 16.81 & 22s \\
MARCO  & \textbf{7.76} & 3s & \textbf{10.72} & 11s & 16.78 & 21s \\
\bottomrule
\end{tabular}
\caption{TSP performance table. The best results overall and the best results among learning-based methods are highlighted in bold.}
\label{table:tsp_perf}
\end{table}


\subsubsection{Generalization to Larger Sizes}

Training NCO models with reinforcement learning is computationally intensive, leading to a common practice in the literature where models are often trained on smaller-sized instances (up to 100). While this approach is understandable due to resource constraints, it is important to consider the ability of these models to generalize to larger instances. This aspect is crucial for their applicability in real-world scenarios, where problem sizes can vary significantly.

The data presented in Table \ref{table:mis_perf} illustrate this point. Here, even a basic greedy constructive method manages to outperform more complex learning-based methods (DGL, LwD, and FlowNet) in the RB800-1200 instance. This observation underlines the importance of using simple heuristics as a sanity check to assess whether advanced models are effectively generalizing to unseen instances or larger sizes.
Similarly, Table \ref{table:tsp_perf} reveals that a simple Nearest Neighbour heuristic is able to surpass POMO, DACT and NeuOPT in instances of 500 cities.
Even though the underlying model of MARCO has been trained on smaller instances (up to 200 for MC and MIS, and up to 100 for TSP), it is able to maintain a good performance in larger graphs with a lower time cost compared to state-of-the-art heuristic solvers.












\section{Limitations and Future Work}

MARCO offers significant advancements in neural combinatorial optimization. However, it has room for improvement.
A primary concern is the uncontrolled growth of its memory during the optimization process, as it continually stores all the encountered states, leading to increased computational and memory costs. To counter this, future work could focus on implementing mechanisms to prune the memory by removing redundant information. 

Another limitation is the substantial resource requirement for storing entire edge-based solutions in memory (like in TSP). This approach, particularly for large instances, can result in high memory consumption and slower retrieval processes. A promising direction would be to represent solutions in a lower-dimensional space using fixed-size embeddings, effectively reducing the memory footprint while preserving (or even incorporating) necessary information. 

In terms of data retrieval, MARCO currently employs a method based on a weighted average of similarity, which may not fully capture the relationships between solution pairs. A more advanced alternative to consider is the implementation of an attention-based search mechanism. This method would not only prioritize the significance of various stored solutions but could also incorporate the objective values or other distinct characteristics of these solutions to compute their relevance.

Additionally, while not a limitation, applying MARCO to new problems or integrating it with different NCO methods requires careful consideration in how memory information is aggregated with instance features. The nature of the data stored and retrieved can vary significantly depending on the specific problem being addressed.

\section{Conclusion}

In this paper, we have introduced the Memory-Augmented Reinforcement for Combinatorial Optimization (MARCO), a framework for Neural Combinatorial Optimization methods that employs a memory module to store and retrieve relevant historical data throughout the search process. The experiments conducted in the maximum cut, maximum independent set and travelling salesman problems validate MARCO's ability to quickly find high-quality solutions, outperforming or matching the state-of-the-art learning methods.
Furthermore, we have demonstrated that the use of a collaborative parallel-thread scheme contributes to the performance of the model while reducing the computation cost.

\section*{Acknowledgments}
Andoni Irazusta Garmendia acknowledges a predoctoral grant from the Basque Government (ref. PRE\_2020\_1\_0023). This work has been partially supported by the Research Groups 2022-2025 (IT1504-22), the Elkartek Program (KK- 2021/00065,
KK-2022/00106) from the Basque Government. 





\appendix

\section{Further implementation details}

In order to solve the Maximum Cut (MC), Maximum Independent Set (MIS) and Travelling Salesman Problem (TSP), MARCO receives information from three sources: the static instance information, the dynamic (partial) solution and the historical memory data.

\subsection{MC and MIS}
In MC and MIS, the instance information is given by binary edge features $\mathbf{y} \in \mathbb{R}^{|E|}$, which denote the existence of an edge in the set of edges $E$ of the graph $G$. The solution is also encoded in binary node features $\mathbf{x} \in \mathbb{R}^{|V|}$, indicating which of the two sets the node belongs to, and where $V$ is the set of nodes. Visited solutions and performed actions (node flips) are stored in memory. The model retrieves a weighted average of the actions performed in similar solutions and this information is concatenated as an additional node feature.

\subsection{TSP}
In TSP, we follow prior works~\cite{kool2018attention,kwon2020pomo} with some modifications:
\begin{itemize}
    \item While prior works only use node features (city coordinates), we also consider the use of distances between cities as edge features, since we believe the relative information is advantageous for the model.
    \item We implemented a Graph Transformer (GT)~\cite{dwivedi2020generalization} encoder layer that also considers the edge features.
    \item We increased the model size (embedding dimension). We obtained better results scaling it. See the selected model's hyperparameters in Table \ref{hyperparams} (lines 1-5).
    \item ReLU activation function is replaced by SwiGLU \cite{shazeer2020glu}.
    \item We implemented a function to compute an arbitrary number of data augmentations (coordinate rotations).
\end{itemize}

The memory data is aggregated as an edge feature, and gives information about the number of times each edge has been considered in previous solutions or routes. These edge features are used in the MHA mechanism of the decoder~\cite{kool2018attention,kwon2020pomo}, where a linear projection of the memory features is added to the attentions weights.
\begin{equation}
   \text{Attn}(Q, K, V) = \text{softmax}\left(\frac{QK^T}{\sqrt{d_k}} + \mathbf{E}\right) V 
\end{equation}

\subsection{Training in MARCO}
The training hyperparameters can be found in Table \ref{hyperparams} (lines 6-15). We trained a unique model for each problem. Each epoch of training comprised 1000 episodes. Within each episode, 128 new instances of an arbitrary size (50-200 for MC and MIS and 20-100 for TSP) were randomly generated. The optimization of model parameters was conducted using the AdamW optimizer~\cite{loshchilov2017decoupled}, and we clipped the gradient norm at a value of 1.0.

The training of \textit{improvement models} (for MC and MIS) requires additional hyperparameters, such as the discount factor and episode length. 
This necessity comes from the reward mechanism employed in these models, which is based on the concept of discounted future rewards, i.e., focusing on maximizing future gains.  A discount factor of 0.95 and an episode length of 20 steps were selected.

We adopted the following two-phase training pipeline for \textit{constructive models}. The initial phase comprises 200 epochs of training without the incorporation of memory, following the training of POMO~\cite{kwon2020pomo}. 

Subsequently, we fix all model weights and introduce a trainable Feed-Forward network into the decoder, which processes the data retrieved from the memory. For the TSP, the processed memory content is incorporated into the edges, and it is used within the Multi-Head Attention mechanism of the decoder in the same fashion as in the GT encoder. 
This second training stage extends for an additional set of 50 epochs. Within each training episode during this stage, the process begins with a deterministic rollout performed with an empty memory. This is followed by executing 5 construction iterations (line 15) using identical problem instances. In these iterations, the model is penalized based on the similarity of its proposed solutions to those generated in previous iterations.

\subsection{Inference in MARCO}
The performance results discussed in the paper have been obtained using models trained with the hyperparameter values shown in Table \ref{hyperparams} (16-21). 

Lines 16-18 have common parameters for both improvement and constructive methods: the number ($K$) of neighbors to consider in the memory retrieval process, the number of parallel execution threads, and the maximum number of solutions stored in memory (once this number is reached the oldest solutions are replaced).

The \textit{Max steps} parameter denotes the number of improvement steps, while \textit{Constructions} denote the number of solutions built in constructive methods.

Lastly, for the TSP, we modulate the frequency of data retrieval from the memory. Instead of conducting this search at every construction step, we establish a predetermined retrieval frequency (\textit{retrieval frequency}) for performing it, significantly accelerating the process, while having minimal impact on performance due to the marginal changes to the partial solution in consequent steps.


\begin{table}
 \caption{Hyperparameters used in MARCO.}
\label{hyperparams}
  \centering
\begin{tabular}{llrrrr}
\toprule

&Hyperparameter  & MC & MIS  & TSP \\ \midrule
 & MODEL HP & & & \\
1 & Embedding dim &  64 & 64 & 512 \\
2 & Encoding layers & 3 & 3 & 6 \\
3 & Attention heads & 8 & 8 & 16 \\
4 & FF hidden dim & 512 & 512 & 2048 \\
5 & Tanh clipping &  10 & 10 & 10 \\
 &  & & & \\
 & TRAINING HP & & & \\
6 & Instance sizes & 50-200 & 50-200 & 20-100 \\
7 & Learning rate & 1e-4 & 1e-4 & 1e-4 \\
8 & Batch size & 128 & 128 & 128 \\
9 & Number of epochs & 100 & 100 & 250 \\ 
10 & Number of episodes & 1000 & 1000 & 1000 \\ 
11 & Repetition penalty & 1.0 & 0.01 & 0.1 \\
12 & Gradient clipping  & 1 &  1 & 1 \\
13 & Discount factor & 0.95 & 0.95 & - \\
14 & Episode length & 20 & 20 & - \\
15 & Constructions & - & - & 5 \\
 &  & & & \\
 & INFERENCE HP & & & \\
16 & k in retrieval & 20 & 20 & 3 \\
17 & Num of threads & 50 & 50 & 100 \\
18 & Max memory size & $10^5$ & $10^5$ & $10^5$ \\
19 & Max steps & $2 \times N$ & $2 \times N$ & - \\
20 & Constructions & - & - & 10 \\
21 & Retrieval frequency & - & - & 10 \\

\end{tabular}
\end{table}%


\section{Impact of k on MARCO's Performance}

\begin{figure}[t]
\centering
\includegraphics[width=0.99\columnwidth]{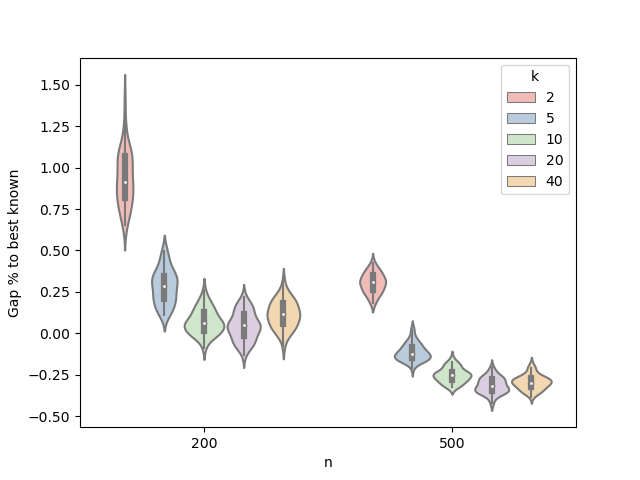}
\caption{Performance of MARCO in MC with different $k$ nearest neighbors retrieved from memory for graphs with 200 and 500 nodes.}
\label{fig:k}
\end{figure}

The number of neighbors, denoted as $k$, considered during the $k$ nearest neighbor search, plays a crucial role in MARCO's performance. As depicted in Figure \ref{fig:k}, small values of $k$ result in poor performance due to the restricted information that can be retrieved at each step. On the other hand, when $k$ is large, the algorithm takes into account numerous neighbors, including those with lower similarity. However, the influence of each retrieved neighbor is mitigated by its similarity, effectively addressing this issue.

\section{Impact of Repetition Penalty Coefficient}
\begin{figure}[t]
\centering
\includegraphics[width=0.99\columnwidth]{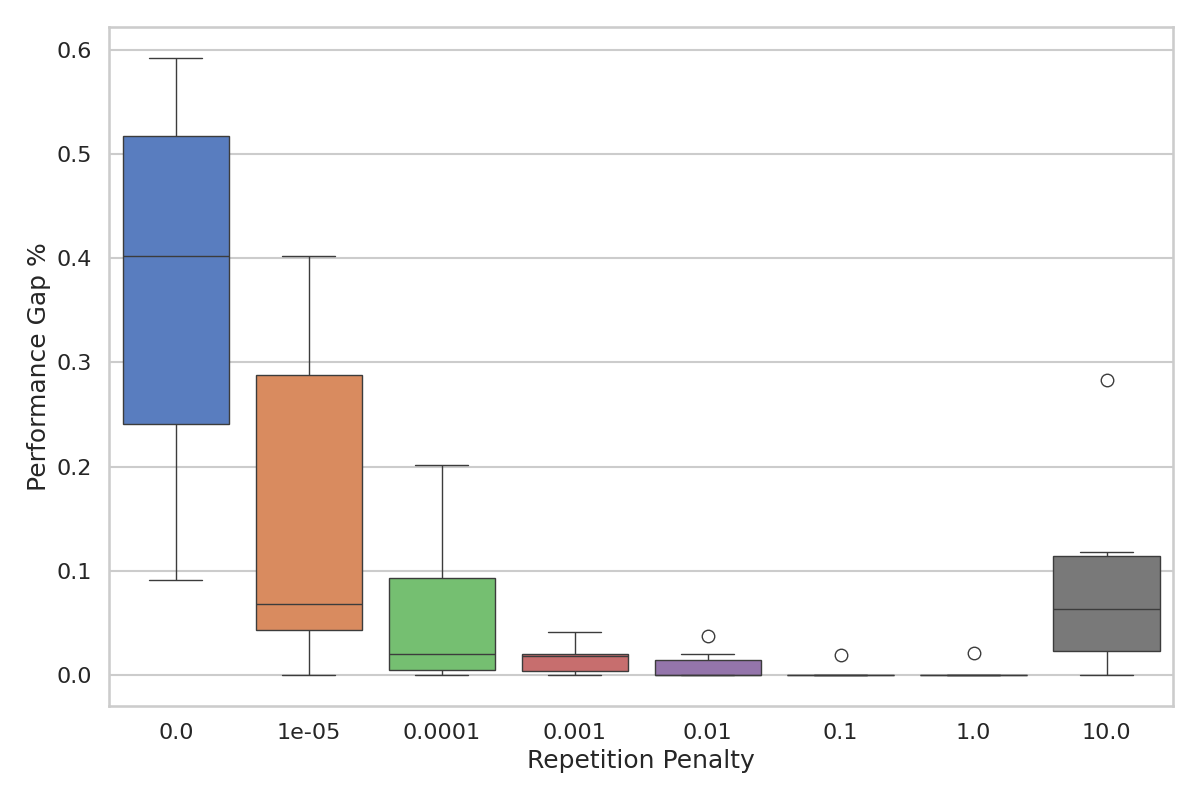}
\caption{Performance of MARCO in MC based on the repetition penalty in the reward used during training.}
\label{fig:punish}
\end{figure}

The penalty coefficient, as indicated in line 11 of Table \ref{hyperparams}, specifies the magnitude of the penalty applied to the reward mechanism when the algorithm suggests a repetitive action. This parameter requires careful adjustment to ensure it harmonizes with the problem's objective value in the reward function. A penalty that is too small may lead the algorithm to frequently repeat actions, whereas an excessively high penalty can generate very large gradient values. Figure \ref{fig:punish} illustrates the results of experiments conducted with various penalty coefficients for the MC problem. We trained five distinct models for each penalty coefficient, following the methodology of the main experiments, and conducting evaluations across 100 graph instances with 200 nodes. The findings reveal that penalty coefficients of 0.1 and 1.0 yield the most favorable performance outcomes.

\section{Analysis of the Memory Module}
\subsection{Theoretical Memory Complexity Analysis}

The Graph Transformer model (GT) and the memory module are the two core elements with the higher memory consumption.

Let us denote $N=|V|$ as the number of nodes in the instance graph. The memory complexity of \textit{transformers}~\cite{vaswani2017attention}, and by extension the GT, is known to scale quadratically with $N$ ($\mathcal{O}(N^2)$).

The complexity associated with the \textit{memory module} varies based on the method employed for storing solutions encountered during the optimization process:
\begin{itemize}
    \item \textit{Node-wise storage}. In MC and MIS, the solutions are encoded in nodes, represented by a vector of length $N$.
    \item \textit{Edge-wise storage}. In TSP, each solution is represented by a vector of size proportional to the number of edges, which in a dense graph is of length $N^2$.
\end{itemize}

Additionally, with each optimization step $t$, a new solution is added to the memory. Consequently, if the optimization process runs for $T$ steps, the memory requirement for storing these solutions scales linearly with $T$.

Therefore, the overall memory complexity of MARCO, depending on the storage method, can be summarized as:
\begin{itemize}
    \item \textit{Node-wise storage}: $\mathcal{O}(N^2 + T \cdot N )$
    \item \textit{Edge-wise storage}: $\mathcal{O}(N^2 + T \cdot N^2)$
\end{itemize}
In scenarios where $T$ is considerably large, the memory requirement associated with solution storage may become the predominant factor in the system's memory complexity.

\begin{figure*}[htbp]
\centering
\begin{subfigure}[b]{0.48\textwidth}
    \includegraphics[width=\textwidth]{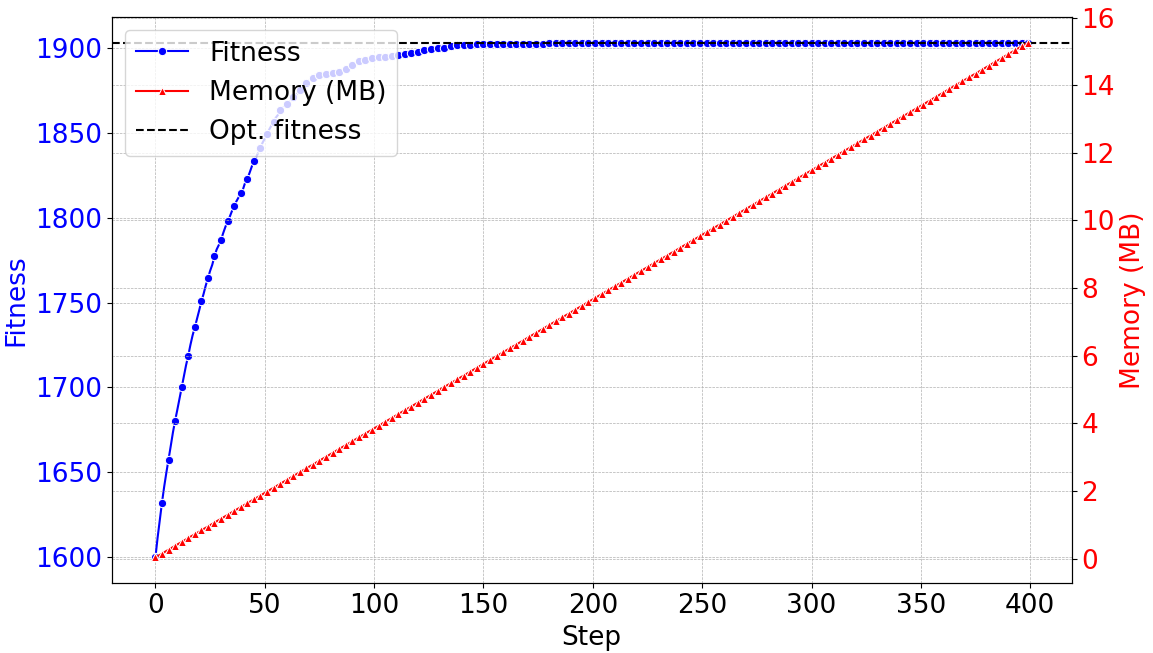}
    \caption{Size 200}
\end{subfigure}
\hfill 
\begin{subfigure}[b]{0.48\textwidth}
    \includegraphics[width=\textwidth]{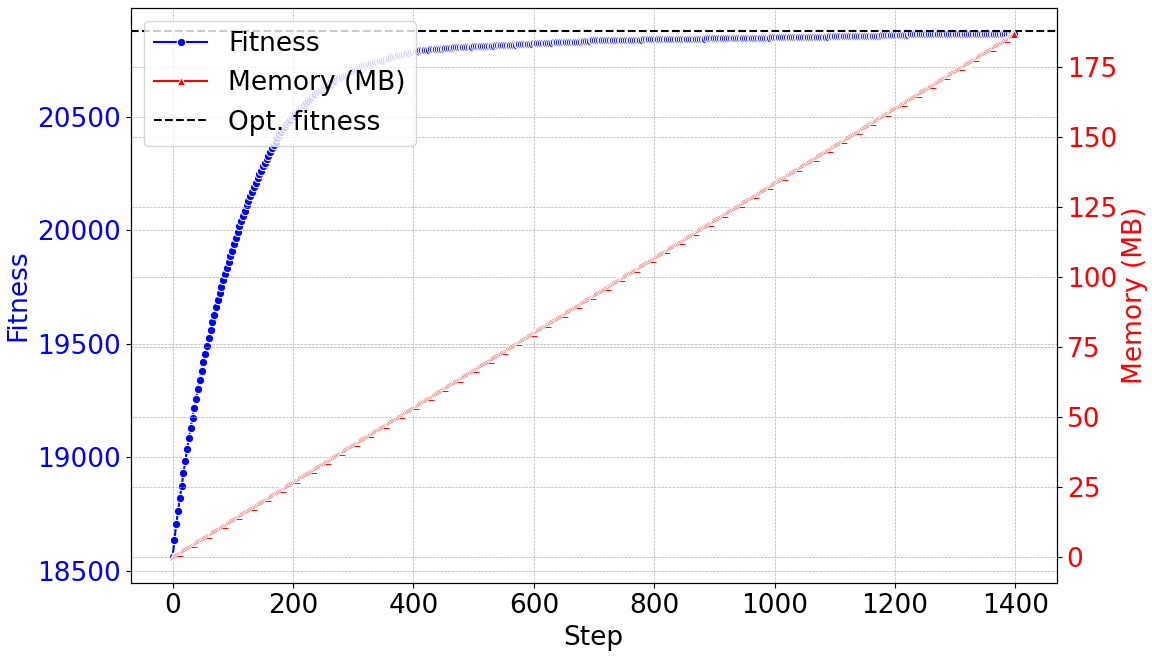}
    \caption{Size 700}
\end{subfigure}
\caption{Performance and Memory Consumption Analysis of MARCO in MC. This graph illustrates the objective value (performance) and its GPU memory usage across successive improving iterations for instances of size (a) 200 and (b) 700.}
\label{fig:perf_mem}
\end{figure*}

\subsection{Empirical Analysis of Memory Usage}

In this section, we provide an empirical analysis of MARCO's memory usage to complement the theoretical analysis. 

Figure \ref{fig:perf_mem} shows the progression of both the objective function and MARCO's memory usage across different sizes of MC instances throughout the optimization process. Consistent with our theoretical findings, the memory usage scales linearly with the number of optimization steps $T$, given a fixed instance size. Notably, the observed memory requirements for achieving (near) optimal solutions are relatively modest, especially when compared to the memory capacities of modern GPUs.

It is important to note that in our implementation for the TSP, where we store the solutions using graph edges (vectors of length $N^2$ rather than $N$), memory consumption is higher. To evaluate the feasibility of running MARCO with this edge-based representation in TSP, we analyzed its memory usage. Our findings reveal that for TSP instances with 100, 200, and 500 cities, our implementation of MARCO required an average memory of 0.04 MB, 0.15 MB, and 0.9 MB, respectively, for storing each construction.

Given that a modern GPU is typically equipped with more than 8GB of memory, the number of possible constructions for the largest TSP size evaluated (500 cities) stands at approximately $\frac{8 \times 1024}{0.9} \thickapprox 9000$. However, in scenarios where MARCO might be required to run for extended periods or handle even larger problem instances, strategies for optimizing memory usage, like those discussed in the future work section of the paper, should be considered.


\bibliographystyle{plain}
\bibliography{ijcai24}

\end{document}